\documentclass[letterpaper, 10 pt, conference]{ieeeconf}  

\IEEEoverridecommandlockouts                              
\usepackage{graphicx}
\usepackage{amsmath} 
\usepackage{amssymb}  

\title{\LARGE \bf
A Survey of Modern Object Detection Literature using Deep Learning
}

\author{Karanbir Chahal$^{1}$ and Kuntal Dey$^{2}$
\thanks{*This work was not supported by any organization}
\thanks{$^{1}$ Karanbir is an independent researcher }
\thanks{$^{2}$ Kuntal Dey is a Member of IBM Academy of Technology (AoT) }}

\begin{document}

\maketitle
\thispagestyle{empty}
\pagestyle{empty}


\begin{abstract}
Object detection is the identification of an object
in the image along with its localization and classification. It has wide spread applications and is a critical component for vision based software systems. This paper seeks to perform a rigorous survey of modern object detection algorithms that use deep learning. As part of the survey, the topics explored include various algorithms, quality metrics, speed/size trade offs and training methodologies. This paper focuses on the two types of object detection algorithms- the SSD class of single step detectors and the Faster R-CNN class of two step detectors. Techniques to construct detectors that are portable and fast on low powered devices are also addressed by exploring new lightweight convolutional base architectures. Ultimately, a rigorous review of the strengths and weaknesses of each detector leads us to the present state of the art. 

\end{abstract}

\section{Introduction}
\label{sec:intro}

\subsection{Background and Motivation}
Deep Learning has revolutionized the computing landscape leading to a fundamental change in how applications are being created. Andrej Karpathy, rightfully coined it as Software 2.0. Applications are fast becoming intelligent and capable of performing complex tasks- tasks that were initially thought of being out of reach for a computer. Examples of these complex tasks include detecting and classifying objects in an image, summarizing large amounts of text, answering questions from a passage, generating art and defeating human players at complex games like Go and Chess.
The human brain processes large amounts of data of varying patterns. It
identifies these patterns, reasons about them and takes some action
specific to that pattern.
Artificial Intelligence aims to replicate this approach through Deep Learning. Deep Learning
has proven to have been quite instrumental in understanding data of
varying patterns at an accurate rate. This capability is responsible for
most of the innovations in understanding language and images.
With Deep Learning research moving forward at a fast pace, new
discoveries and algorithms have led to disruption of numerous fields.
One such field that has been affected by Deep Learning in a substantial
way is object detection.
\subsection{Object Detection}
Object detection is the identification of an object
in an image along with its localization and classification. Software
systems that can perform these tasks are called object detectors.
Object Detection has important applications. Numerous tasks
which require human supervision can be automated with a software
system that can detect objects in images. These include surveillance,
disease identification and driving. The advent of deep learning has brought a profound change in how we implement computer vision nowadays. \cite{c22}

Unfortunately, this technology has a high potential for irresponsible use. Military applications of object detectors are particularly worrying. Hence, in spite of its considerable useful applications, caution and responsible usage should always be kept in mind.
\subsection{Progress and Future Work}
Object Detectors have been making fast strides in accuracy, speed and
memory footprint. The field has come a long way since 2015, when the first viable deep learning based
object detector was introduced. The earliest deep learning object detector took $47s$ to process
an image, now it takes less than $30ms$ which is better than real time. Similar to
speed, accuracy has also steadily improved.
From a detection accuracy of $29$ mAP (mixed average precision),
modern object detectors have achieved $43$ mAP.
Object detectors have also improved upon their size. Detectors can run well on low powered phones, thanks to the intelligent and conservative
design of the models. Support for running models on phones has  improved thanks to frameworks like Tensorflow\cite{c34} and Caffe\cite{c35} among others.
A decent argument can be made that object detectors have achieved
close to human parity. Conversely, like any deep
learning model, these detectors are still open to adversarial attacks and
can misclassify objects if the image is adversarial in nature.
Work is being done to make object detectors and deep learning models
in general more robust to these attacks. Accuracy, speed and size will
constantly be improved upon, but that is no longer the most pressing
goal. Detectors have attained a respectable quality, allowing them to be put into production today. The goal now should be to make these
models robust against hacks and ensure that this technology is being
used responsibly.

\section{Object Detection Models}
\label{sec:models}

\subsection{Early Work}

The first object detector came out in 2001 and was called the Viola Jones Object Detector \cite{c7}. Although, it was technically classified as an object detector, it's primary use case was for facial detection. It provided a real time solution and was adopted by many computer vision libraries at the time. The field was substantially accelerated with the advent of Deep Learning. The first Deep Learning object detector model was called the Overfeat Network \cite{c13} which used Convolutional Neural Networks (CNNs) along with a sliding window approach. It classified each part of the image as an object/non object and subsequently combined the results to generate the final set of predictions. This method of using CNNs to solve detection led to new networks being introduced which pushed the state of the art even further. We shall explore these networks in the next section.

\subsection{Recent Work}
There are currently two methods of constructing object detectors- the
single step approach and the two step approach. The two step approach
has achieved a better accuracy than the former whereas the single step approach has been faster and shown higher memory efficiency.
The single step approach classifies objects in images along with their locations in a single step.
The two step approach on the other hand divides this process into two steps. The first step
generates a set of regions in the image that have a high probability of being an
object. The second step then performs the final detection and
classification of objects by taking these regions as input. These two
steps are named the \textit{Region Proposal Step} and the \textit{Object Detection
Step} respectively.
Alternatively, the single step approach combines these two steps to directly predict the class
probabilities and object locations.

Object detector models have gone through various changes throughout
the years since 2012. The first breakthrough in object detection was the RCNN \cite{c1} which resulted in an improvement of nearly 30\% over the previous state of the art. We shall start the survey by exploring this detector first.

\subsection{Problem Statement for Object Detection}
There are always two components to constructing a deep learning
model. The first component is responsible for dividing the training data
into input and targets. The second component is deciding upon the
neural network architecture and training regime.
The input for these models is an image. The targets are a list of object
classes relaying what class the object belongs to and their
corresponding coordinates. These coordinates signify where in the
image the object exist. There are 4 types of coordinates- the center x
and y coordinates and the height and width of the bounding box. We shall use the term bounding box to denote the box formed by applying these 4 coordinates on the image. The network is trained to predict a list of objects with their corresponding
locations in the form of bounding box coordinates.

\section*{Two Step Models}

\section{Region Convolutional Network (R-CNN)}
\label{sec:rcnn}

\begin{figure*}[htb]
\centering
\includegraphics[width=0.95\textwidth]{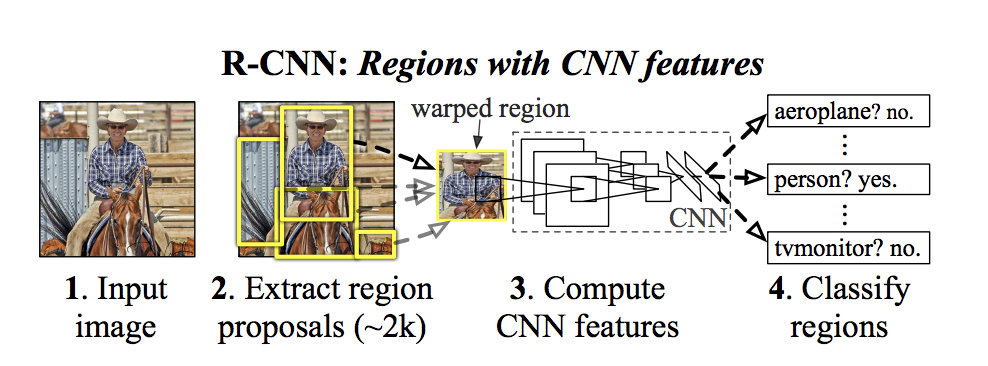}
\label{fig:model1}
\caption{Region Convolutional Network}
\end{figure*} 

The RCNN Model \cite{c1} was a highly influential model that has shaped the structure of modern object detectors. It was the first detector which proposed the two step approach. We shall first look at the Region Proposal Model now.
\subsection{Region Proposal Model}

In this model, the image is the input. A region proposal system finds a set of blobs or regions in the image which have a high
degree of probability of being objects.
The Region Proposal System for the R-CNN uses a non deep learning
model called Selective Search \cite{c12}. Selective Search finds a list of regions that it deems most plausible of having an object in them. It  finds a large number of regions  which are then cropped from the input image and resized to a size of 7 by 7 pixels. These blobs are then fed into the
\textit{Object Detector Model}. The Selective Search outputs around ~2,000 region proposals of various scales and takes approximately ~27 seconds to execute.

\subsection{Object Detector Model}

Each deep learning model is broken down into 5 subsections in  this paper. These are the \textit{input} to the model, the \textit{targets} for the model to learn on, the \textit{architecture}, the \textit{loss function} and the \textit{training procedure} used to train the model.

\begin{itemize}

\item \textbf{Input}: The object detector model takes the 7 by 7  sized regions calculated by the region proposal model as input.

\item \textbf{Targets}: The targets for the RCNN network are a list of class probabilities and offset coordinates for each region proposed by the Selective Search. The box coordinate offsets are calculated to allow the network to learn to fit objects better. In other words, offsets are used to modify the original shape of the bounding box such that it encloses the object exactly. The region is assigned a class by calculating the Intersection Over Union (IOU) between it and the ground truth. If the IOU $\geq$ 0.7, the region is assigned the class. If multiple ground truth’s have an IOU $\geq$  0.7 with the region, the ground truth with the highest IOU is assigned to that region. If the IOU $\leq$  0.3, the region is assigned the background class. The other regions are not used in calculation of the loss and are hence ignored during training. The offsets between the ground truth and that region are calculated only if a region is allotted a foreground class. The method for calculating these offsets varies. In the RCNN series, they are calculated as follows:

$$
t_x = (x_g - x_a)/w_a  \eqno{(1)}
$$
$$
t_y = (y_g - y_a)/h_a  \eqno{(2)}
$$
$$
t_w = log(w_g/w_a)  \eqno{(3)}
$$
$$
t_h = log(h_g/h_a)  \eqno{(4)}
$$

where $x_g$, $y_g$, $w_g$ and $h_g$ are the x,y coordinates, width and height of the ground truth box and $x_a$, $y_a$, $w_a$ and $h_a$ are the x,y coordinates, width and height of the region. The four values $t_x$, $t_y$, $t_w$ and $t_h$ are the target offsets.

\item \textbf{Architecture}: The architecture of the Object Detector Model consists of a series of convolutional and max pooling layers with activation functions. A region from the above step is run through these layers to generate a feature map. This feature map denotes a high level format of the region that is interpretable by the model. The feature map is unrolled and fed into two fully connected layers to generate a 4,078 dimensional vector. This vector is then input into two separate small SVM networks \cite{c41} - the \textit{classification network head} and the \textit{regression network head}.
The classification head is responsible for predicting the class of object that the region belongs to and the regression head is responsible for predicting the offsets to the coordinates of the region to better fit the object.
\item \textbf{Loss}: There are two losses computed in the RCNN- the classification loss and the regression loss. The classification loss is the cross entropy loss and the regression loss is an L2 loss.
The RCNN uses a multi step training pipeline to train the network using these two losses.

The cross entropy loss is given as:
$$
CE({\boldsymbol y}, \hat{\boldsymbol y}) = - \sum^{\textrm{N}_{c}}_{i=1} y_{i}\log(\hat{y}_{i}) \eqno{(5)}
$$
where ${\boldsymbol y}\in\mathbb{R}^{5}$ is a one-hot label vector and $\textrm{N}_{c}$ is the number of classes.

The L2 or the Mean Square Error (MSE) Loss is given as :

$$
L2(x_1,x_2) = \frac{1}{n} \left \| x_1-x_2 \right \|_2^2 = \frac{1}{n} \sum_i (x_{1_i}-x_{2_i})^2 \eqno{(6)}
$$

\item \textbf{Model Training Procedure}: The model is trained using a two step procedure. Before training, a convolutional base pre-trained on ImageNet is used.
The first step includes training the SVM classification head using the cross entropy loss. The weights for the regression head are not updated.
In the second step, the regression head is trained with the L2 loss. The weights for the classification head are fixed.
This process takes approximately 84 hours as features are computed and stored for each region proposal. The high number of regions occupy a large amount of space and the input/output operations add a substantial overhead.
\item \textbf{Salient Features}:  The RCNN model takes \textit{47 seconds} to process a single image since it has a complex multistep training pipeline which requires careful tweaking of parameters. Training is expensive in both time and space. The features computed for the dataset occupy hundreds of gigabytes and take around 84 hours to train. The RCNN provided a good base to iterate upon by providing a structure to solve the object detection problem. However, due to its time and space constraints, a better model was needed.
\end{itemize}

\section{Fast RCNN}

\label{sec:fast-rcnn}

\begin{figure*}[htb]
\centering
\includegraphics[width=0.95\textwidth]{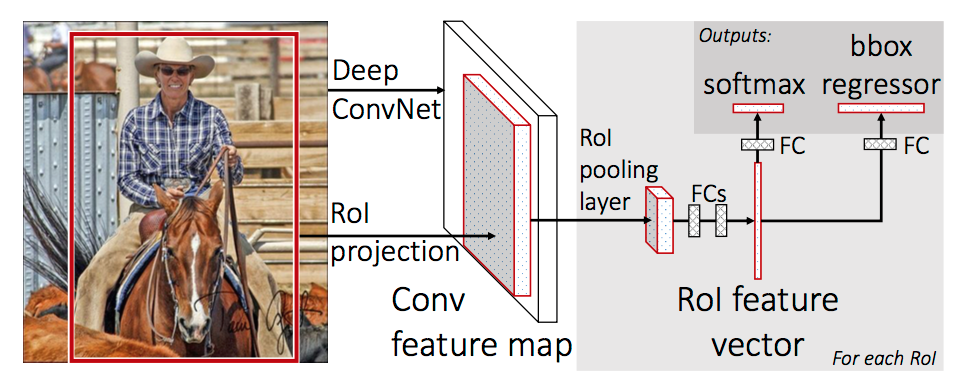}
\label{fig:model2}
\caption{Fast RCNN}
\end{figure*} 

The Fast RCNN \cite{c3} came out soon after the RCNN and was a substantial improvement upon the original. The Fast RCNN is also a two step model which is quite similar to the RCNN, in that it uses selective search to find some regions and then runs each region through the object detector network. This network consists of a convolutional base and two SVM heads for classification and regression. Predictions are made for the class and offsets of each region.
The RCNN Model takes every region proposal and runs them through the convolutional base. This is quite inefficient as an overhead of running a region proposal through the convolutional base is added, everytime a region proposal is processed. The Fast RCNN aims to reduce this overhead by running the convolutional base just once. It runs the convolutional base over the entire image to generate a feature map. The regions are cropped from this feature map instead of the input image. Hence, features are shared leading to a reduction in both space and time. This cropping procedure is done using a new algorithm called ROI Pooling.

The Fast RCNN also introduced a single step training pipeline and a multitask loss, enabling the classification and regression heads to be trained simultaneously. These changes led to substantial decrease in training time and memory needed. 
Fast RCNN is more of a speed improvement than an accuracy improvement. It takes the ideas of RCNN and packages it in a more compact architecture.
The improvements in the model included a single step end-to-end training pipeline instead of a multi step pipeline and reduced training time from 84 hours to \textit{9 hours}. It also had a reduced memory footprint, no longer requiring features to be stored on disk.
The major innovation of Fast RCNN was in sharing the features of a convolutional net. Constructing a single step training pipeline instead of a multistep training pipeline by using a multitask loss was also a novel and elegant solution.

\subsection{ROI Pooling} 
Region of Interest (ROI) Pooling is an algorithm that takes the coordinates of the regions obtained via the Selective Search and directly crops it out from the feature map of the original image. ROI Pooling allows for computation to be shared for all regions as the convolutional base need not be run for each region. The convolutional base is run only once for the input image to generate a single feature map. Features for various regions are computed by \textit{cropping} this feature map. 

In the ROI Pooling algorithm, the coordinates of the regions proposed are divided by a factor of $h$, the compression factor. The compression factor is the amount by which the image is compressed after it is run through the convolutional base. The value for $h$ is 16 if the VGGNet \cite{c47} is used as the convolutional base. This value was chosen because the VGGNet compresses the image to $1/16^{th}$ of its original width and height.

The compressed coordinates are calculated as follows: 

$$
x_{new} = x_{old}/h
$$
$$
y_{new} = y_{old}/h
$$
$$
w_{new} = w_{old}/h
$$
$$
h_{new} = h_{old}/h
$$

where $x_{new}$, $y_{new}$, $w_{new}$ and $h_{new}$ are the compressed x,y coordinates, width and height.

Once the compressed coordinates are calculated, they are plotted on the image feature map. The region is cropped and resized from the feature map to a size of 7 by 7. This resizing is done using various methods in practice. In ROI Pooling, the region plotted on the feature map is divided into 7 by 7 bins. Max pooling is performed on the cells in each bin. Often in practice, a variation of ROI Pooling called ROI Averaging is used. It simply replaces the max pooling with average pooling. The procedure to divide this feature map into 49 bins is approximate. It is not uncommon for one bin to contain more number of cells than the other bins.

Some object detectors simply resize the cropped region from the feature map to a size of 7 by 7 using common image algorithms instead of going through the ROI Pooling step. In practice, accuracy isn't affected substantially by doing this. Once these regions are cropped, they are ready to be input into the object detector model.

\subsection{Object Detector Model}

The object detector model in the Fast RCNN is very similar to the detector used in the RCNN.

\begin{itemize} 
\item  \textbf{Input}: The object detector model takes in the region proposals received from the region proposal model. These region proposals are cropped using ROI Pooling.

\item  \textbf{Targets}: The class targets and box regression offsets are calculated in the exact same way as done in the RCNN.
\item  \textbf{Architecture}: After the ROI Pooling step, the cropped regions are run through a small convolutional network. This network consists of a set of convolutional and fully connected layers. These layers output a 4,078 dimensional vector which is in turn used as input for the classification and regression SVM heads for class and offsets predictions respectively.

\item  \textbf{Loss}: The model uses a multitask loss which is given as:
$$
L = l_c + \alpha*\lambda*l_r  \eqno{(7)}
$$
where $\alpha$ and $\lambda$ are hyperparameters.
The $\alpha$ hyperparameter is switched to 1 if the region was classified with a foreground class and 0 if the region was classified as a background class. The intuition is that the loss generated via the regression head should only be taken into account if the region actually has an object in it. The 
$\lambda$  hyperparameter is a weighting factor which controls the weight given to each of these losses. It is set to 1 in training the network. This loss enables joint training.
	The loss of classification ($l_c$) is a regular log loss and the loss of regression ($l_r$) is a Smooth $L_1$ loss. The Smooth $L_1$ loss is an improvement over the $L_2$ loss used for regression in RCNN. It is found to be less sensitive to outliers as training with unbounded regression targets leads to gradient explosion. Hence, a carefully tuned learning rate needs to be followed. Using the Smooth $L_1$ loss removes this problem.

The Smooth $L_1$ loss is given as follows:

$$
  \label{smoothL1}
  \textrm{Smooth}_{L_1}(x) =
  \begin{cases}
  0.5x^2& \text{if } |x| < 1\\
  |x| - 0.5& \text{otherwise}
  \end{cases}
  \eqno{(8)}
$$
This is a robust $L_1$ loss that is less sensitive to outliers than the $L_2$ loss used in R-CNN.

\item  \textbf{Model Training Procedure}: In the RCNN, training had two distinct steps. The Fast RCNN introduces a single step training pipeline where the classification and regression subnetworks can be trained together using the multitask loss described above. The network is trained with Synchronous Gradient Descent (SGD) with a mini batch size of 2 images. 64 random region proposals are taken from each image resulting in a mini batch size of 128 region proposals.

\item  \textbf{Salient Features}: 
Firstly, Fast RCNN shares computation through the ROI Pooling step hence leading to dramatic increases in speed and memory efficiency. More specifically, it reduces training time exponentially from 84 hrs to 9 hrs and also reduces inference time from 47 seconds to 0.32 seconds.  Secondly, it introduces a simpler single step training pipeline and a new loss function. This loss function is easier to train and does not suffer with the gradient explosion problem.

The objector detector step reports real time speeds. However, the region proposal step proves to be a bottleneck. More specifically, a better solution than Selective Search was needed as it was deemed too computationally demanding for a real time system. The next model aimed to do just that. 

\end{itemize}

\section{Faster RCNN}
\label{sec:faster-rcnn}

   \begin{figure}[thpb]
      \centering
      \includegraphics[scale=0.5]{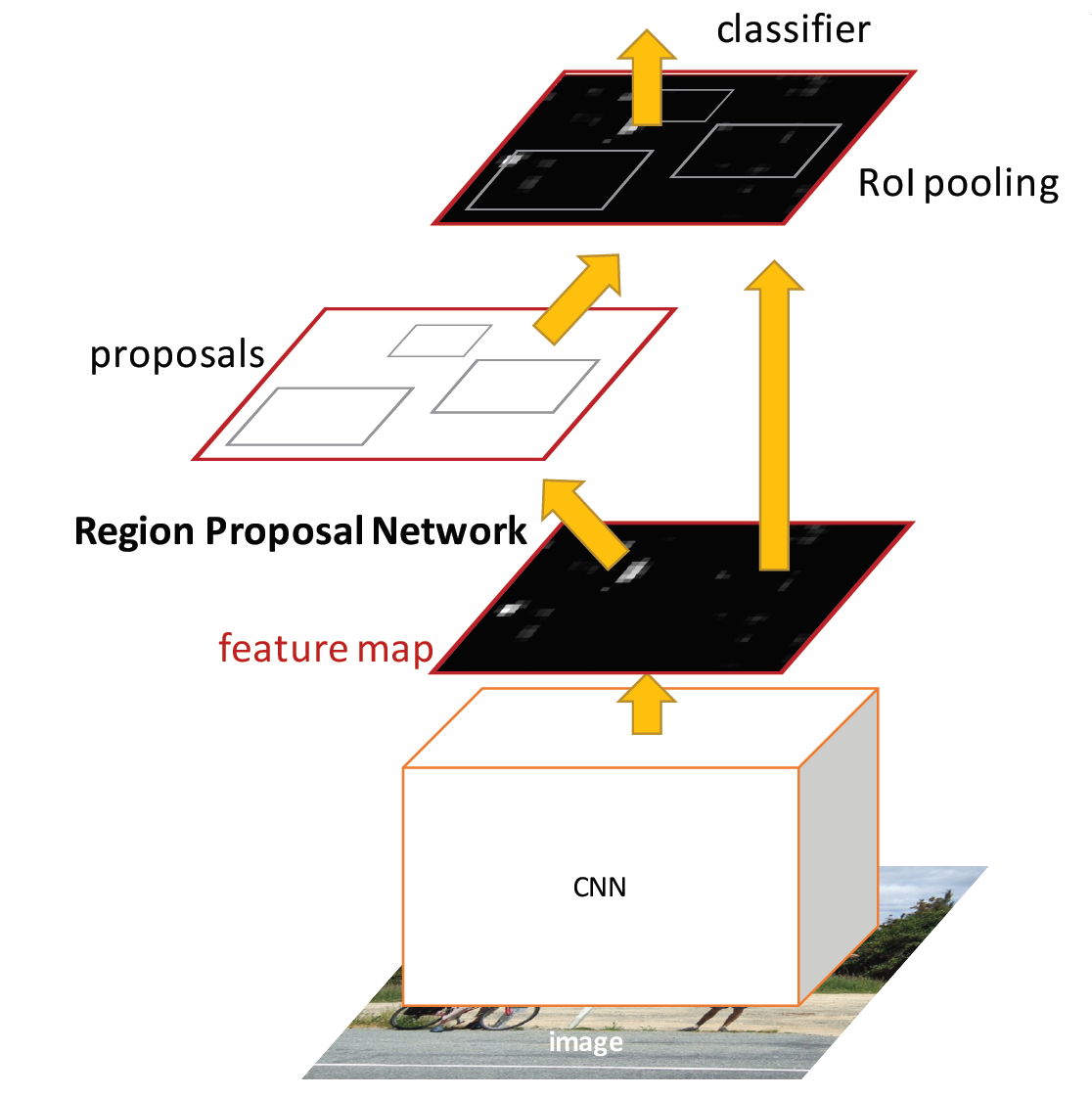}
      \caption{Faster RCNN}
      \label{fig:model3}
   \end{figure}
   

The Faster RCNN \cite{c6} came out soon after the Fast RCNN paper. It was meant to represent the final stage of what the RCNN set out to do. It proposed a detector that was learnt end to end. This entailed doing away with the algorithmic region proposal selection method and constructing a network that learned to predict good region proposals. Selective Search was serviceable but took a lot of time and set a bottleneck for accuracy. A network that learnt to predict higher quality regions would theoretically have higher quality predictions.

The Faster RCNN introduced the \textit{Region Proposal Network} (RPN) to replace Selective Search. The RPN needed to have the capability of predicting regions of multiple scales and aspect ratios across the image. This was achieved using a novel concept of anchors.

\subsection{Anchors}
Anchors are a set of regions in an image of a predefined shape and size i.e anchors are simply rectangular crops of an image. To model for objects of all shapes and sizes, they have a diverse set of dimensions. These multiple shapes are decided by coming up with a set of aspect ratios and scales. The authors use scales of {32px, 64px, 128px} and aspect ratios of {1:1, 1:2, 2:1} resulting in 9 types of anchors. Once a location in the image is decided upon, these 9 anchors are cropped from that location. 	

Anchors are cropped out after every $x$ number of pixels in the image. This process starts at the top left and ends at the bottom right of the image. The window slides from left to right, $x$ pixels at a time, moving $x$ pixels towards the bottom after each horizontal scan is done. The process is also called a sliding window.

The number of pixels after which the set of anchors are cropped out are decided by the compression factor ($h$) of the feature map described above. In VGGNet, that number is 16. In the Faster RCNN paper, 9 crops of different sizes are cropped out after every 16 pixels (in height and width) in a sliding window fashion across the image. In this way, anchors cover the image quite well.

\subsection{Region Proposal Model}

\begin{itemize}

\item \textbf{Input}: The input to the model is the input image. Images are of a fixed size, 224 by 224 and the training data for the model is augmented by using standard tricks like horizontal flipping. To decrease training time, batches of images are fed into the network. The GPU can parallelize matrix multiplications and can therefore process multiple images at a time.

\item \textbf{Targets}: Once the entire set of anchors are cropped out of the image, two parameters are calculated for each anchor- the class probability target and the box coordinate offset target. The class probability target for each anchor is calculated by taking the IOU of the ground truth with the anchor. If the IOU $\geq$ 0.7, we assign the anchor the class of the ground truth object. If there are multiple ground truths with an IOU $\geq$ 0.7, we take the highest one. If the IOU $\leq$ 0.3, we assign it the background class. If 0.3 $\leq$ IOU $\leq$ 0.7, we fill in 0 values as those anchors will not be considered when the loss is calculated, thereby, not affecting training. If an anchor is allotted a foreground class, offsets between the ground truth and the anchor are calculated. These box offsets targets are calculated to make the network learn how to better fit the ground truth by modifying the shape of the anchor. The offsets are calculated in the same way as in the RCNN model.

\item \textbf{Architecture}: The RPN network consists of a convolutional base which is similar to the one used in the RCNN object detector model. This convolutional base outputs a feature map which is in turn fed into two subnetworks- a classification subnetwork and a regression subnetwork.

The classification and regression head consist of few convolutional layers which generate a classification and regression feature map respectively. The only difference in the architecture of the two networks is the shape of the final feature map. The classification feature map has dimensions $w * h * (k*m)$, where $w$, $h$ and $k*m$ are the width, height and depth. The value of $k$ denotes the number of anchors per pixel point, which in this case is 9 and $m$ represents the number of classes.  The feature map for the regression head has the dimensions of $w * h * (k*4)$. The value of 4 is meant to represent the predictions for the four offset coordinates for each anchor. The cells in the feature map denote the set of pixels out of which the anchors are cropped out of. Each cell has a depth which represents the box regression offsets of each anchor type. Similar to the classification head, the regression head has a few convolutional layers which generate a regression feature map.

\item \textbf{Loss} : The anchors are meant to denote the good region proposals that the object detector model will further classify on.
The model is trained with simple log loss for the classification head and the Smooth L1 loss for the regression head. There is a weighting factor of $\lambda$ that balances the weight of the loss generated by both the heads in a similar way to the Fast RCNN loss. The model doesn't converge when trained on the loss computed across all anchors. The reason for this is that the training set is dominated by foreground/background examples, this problem is also termed class imbalance. To evade this problem, the training set is made by collecting 256 anchors to train on, 128 of these are foreground anchors and the rest are background anchors. Challenges have been encountered keeping this training set balanced making it an active area of research.

\end{itemize}

\subsection{Object detector model}
The object detector model is the same model as the one used in Fast RCNN. The only difference is that the input to the model comes from the proposals generated by the RPN instead of the Selective Search.

\begin{itemize}

\item \textbf{Salient Features}: The Faster RCNN is faster and has end to end deep learning pipeline. The network improved state of the art accuracy due to the introduction of the RPN which improved region proposal quality.

\end{itemize}

\section{Extensions to Faster RCNN}

There have been various extensions to Faster RCNN network that have made it faster and possess greater scale invariance. To make the Faster RCNN scale invariant, the original paper took the input image and resized it to various sizes. These images were then run through the network. This approach wasn't ideal as the network ran through one image multiple times, making the object detector slower. Feature Pyramid Networks provide a robust way to deal with images of different scales while still retaining real time speeds. Another extension to the Faster RCNN framework is the Region Fully Convolutional Network (R-FCN). It refactors the original network by making it fully convolutional thus yielding greater speeds. We shall elaborate on both of these architectures in the coming sections. 

\subsection{Feature Pyramid Networks (FPN)}

\label{sec:fpn}

 \begin{figure}[thpb]
      \centering
      \includegraphics[scale=0.5]{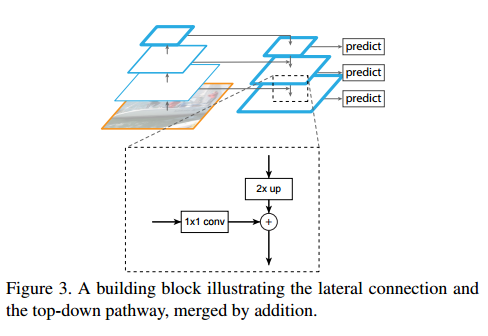}
      \caption{Feature Pyramid Network}
      \label{fig:model4}
   \end{figure}


Scale invariance is an important property of computer vision systems. The system should be able to recognize an object up close and also from far away. The Feature Pyramid Network \cite{c9} (FPN) provides such a neural network architecture.

\begin{figure*}[htb]
\centering
\includegraphics[width=0.95\textwidth scale=1.0]{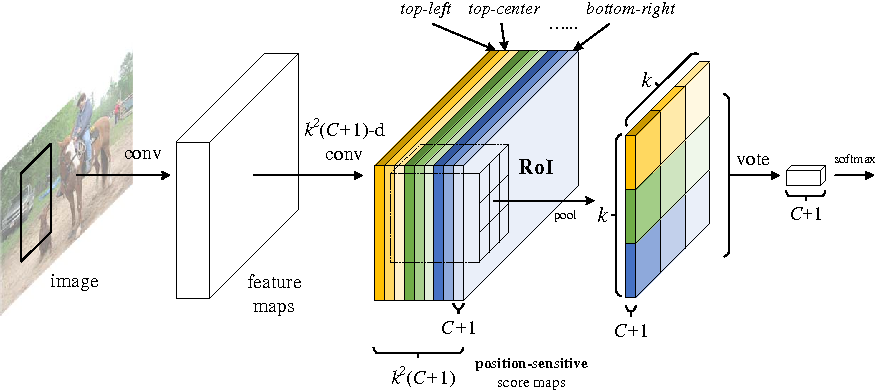}
\label{fig:model5}
\caption{Region - Fully Convolutional Network (R-FCN)}
\end{figure*}

In the original Faster RCNN, a single feature map was created. A classification head and a regression head were attached to this feature map. However, with a FPN there are multiple feature maps that are designed to represent the image at different scales. The regression and classification heads are run across these multi scale feature maps. Anchors no longer have to take care of scale. They can only represent various aspect ratios, as scale is handled by these multi scale feature maps implicitly. 

\begin{itemize}

\item \textbf{Architecture}: The FPN model takes in an input image and runs it through the convolutional base. The convolutional base takes the input through various scales, steadily transforming the image to be smaller in height and width but deeper in channel depth.
This process is also called the \textit{bottom up pathway}. For example, in a ResNet the image goes through five scales which are 224, 56, 28, 14 and 7. This corresponds to four feature maps in the author's version of FPN (the first feature map is ignored as it occupies too much space). These feature maps are responsible for the anchors having scales of 32px, 64px, 128px and 256px. These feature maps are taken from the last layer of each scale- the intuition being that the deepest features contain the most salient information for that scale. Each of the four feature maps goes through a 1 by 1 convolution to bring the channel depth to $C$. The authors used $C = 256$ channels in their implementation. These maps are then added element wise to the upsampled version of the feature map one scale above them. This procedure is also called a \textit{lateral connection}. The upsampling is performed using nearest neighbour sampling with a factor of 2. This upsampling procedure is also called a \textit{top down pathway}. Once lateral connections have been performed for each scale, the updated feature maps go through a 3 by 3 convolution to generate the final set of feature maps. This procedure of lateral connections that merge bottom up pathways and top down pathways, ensures that the feature maps have a high level of information while still retaining the low level localization information for each pixel. It provides a good compromise between getting more salient features while still retaining the overall structure of the image at that scale. 

After generating these multiple feature maps, the Faster RCNN network runs on each scale. Predictions for each scale are generated, the major change being that the regression and classification heads now run on multiple feature maps instead of one. FPN’s allow for scale invariance in testing and training images. Previously, Faster RCNN was trained on multi scaled images but testing on images was done on a single scale. Now, due to the structure of FPN, multi scale testing is done implicitly.

\item \textbf{Salient Points}: New state of the art results were obtained in object detection, segmentation and classification by integrating FPNs into the pre-existing models.

\end{itemize}

\begin{figure*}[htb]
\centering
\includegraphics[width=0.95\textwidth]{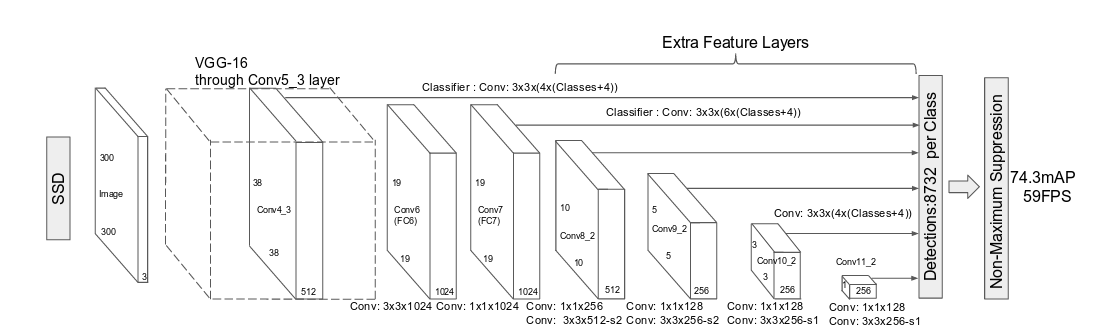}
\label{fig:model6}
\caption{Single Shot MultiBox Detector (SSD)}
\end{figure*} 

\subsection{Region - Fully Convolutional Network (R-FCN)}
\label{sec:rfcn}

In the Faster RCNN after the RPN stage, each region proposal had to be cropped out and resized from the feature map and then fed into the Fast RCNN network. This proved to be the most time consuming step in the model and the research community focused on improving this. The R-FCN is an attempt to make the Faster RCNN network faster by making it fully convolutional and delaying this cropping step.

There are several benefits of a fully convolutional network. One of them is speed- computing convolutions is faster than computing a fully connected layer. The other benefit is that the network becomes scale invariant. Images of various sizes can be input into the network without modifying the architecture because of the absence of a fully connected layer. Fully convolutional networks first gained popularity with segmentation networks \cite{c48}. The R-FCN refactors the Faster RCNN network such that it becomes fully convolutional.

\begin{itemize}

\item \textbf{Architecture}: Instead of cropping each region proposal out of the feature map, the R-FCN model inputs the entire feature map into the regression and classification heads, bringing their depth to a size of $z_r$ and $z_c$ respectively. The value of $z_c$ is $k*k*(x)$, where $k$ is 7, which was the size of the side of the crop after ROI Pooling and $x$ represents the total number of classes. The value of $z_r$ is $k*k*4$, where 4 represents the number of box offsets.

The process of cropping a region is similar to ROI Pooling. However, instead of max pooling the values from each bin on the single feature map, max pooling is performed on different feature maps. These feature maps are chosen depending on the position of the bin. For example, if max pooling is needed to be done for the $i^{th}$ bin out of $k*k$ bins, the $i^{th}$ feature map would be used for each class. The coordinates for that bin would be mapped on the feature map and a single max pooled value will be calculated. Using this method an ROI map is created out of the feature map. The probability value for an ROI is then calculated by simply finding out the average or maximum value for this ROI map. A softmax is computed on the classification head to give the final class probabilities. 

Hence, an R-FCN modifies the ROI Pooling and does it at the end of the convolutional operations. There is no extra convolution layer that a region goes through after the ROI Pooling. The R-FCN shares features in a better way than the Faster RCNN while also reporting speed improvements. It retains the same accuracy as the Faster RCNN.

\item \textbf{Salient Features}: The R-FCN sets a new state of the art in the speed of two step detectors. It achieves an inference speed of 170ms, which is 2.5 to 20x faster than it’s Faster RCNN counterpart.

\end{itemize}

\section*{Single Step Object Detectors}
Single Step Object Detectors have been popular for some time now. Their simplicity and speed coupled with reasonable accuracy have been powerful reasons for their popularity. Single step detectors are similar to the RPN network, however instead of predicting objects/non objects they directly predict object classes and coordinate offsets.

\section{Single Shot MultiBox Detector (SSD)}
\label{sec:ssd}
Single Shot MultiBox Detector \cite{c5} \cite{c18} came out in 2015, boasting state of the art results at the time and real time speeds. The SSD uses anchors to define the number of default regions in an image. As explained before, these anchors predict the class scores and the box coordinates offsets. A backbone convolutional base (VGG16) is used and a multitask loss is computed to train the network. This loss is similar to the Faster RCNN loss function- a smooth L1 loss to predict the box offsets is used along with the cross entropy loss to train for the class probabilities. The major difference between the SSD from other architectures is that it was the first model to propose training on a feature pyramid. 

The network is trained on $n$ number of feature maps, instead of just one. These feature maps, taken from each layer are similar to the FPN network but with one important difference. They do not use top down pathways to enrich the feature map with higher level information. A feature map is taken from each scale and a loss is computed and back propagated. Studies have shown that the top down pathway is important in ablation studies. Modern object detectors modify the original SSD architecture by replacing the SSD feature pyramid with the FPN.
The SSD network computes the anchors for each scale in a unique way. The network  uses a concept of aspect ratios and scales, each cell on the feature map generates 6 types of anchors, similar to the Faster RCNN. These anchors vary in aspect ratio and the scale is captured by the multiple feature maps, in a similar fashion as the FPN. SSD uses this feature pyramid to achieve a high accuracy, while remaining the fastest detector on the market. It's variants are used in production systems today, where there is a need for fast low memory object detectors. Recently, a tweak to the SSD architecture was introduced which further improves on the memory consumption and speed of the model without sacrificing on accuracy. The new network is called the Pyramid Pooling Network \cite{c38}. The PPN replaces the convolution layers needed to compute feature maps with max pooling layers which are faster to compute.

\begin{figure*}[htb]
\centering
\includegraphics[width=0.95\textwidth]{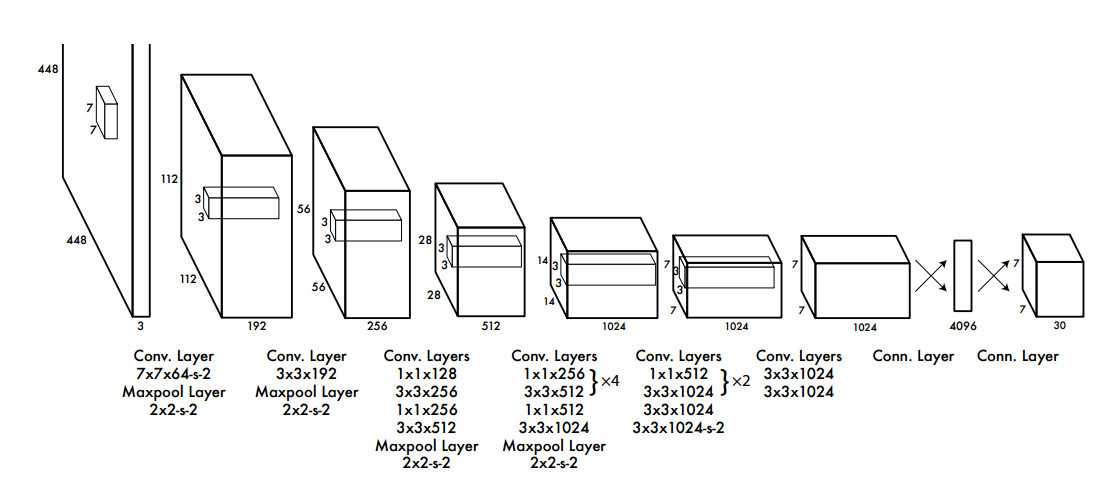}
\label{fig:model7}
\caption{You Only Look Once (YOLO)}
\end{figure*}

\section{You Only Look Once (YOLO)}
\label{sec:yolo}

The YOLO \cite{c10}\cite{c11} group of architectures were constructed in the same vein as the SSD architectures. The image was run through a few convolutional layers to construct a feature map. The concept of anchors was used here too, with every grid cell acting as a pixel point on the original image. The YOLO algorithm generated 2 anchors for each grid cell. Unlike the Fast RCNN, Yolo has only one head. The head outputs feature map of size 7 by 7 by $(x+1+5*(k))$, $k$ is the number of anchors, $x+1$ is the total number of classes including the background class. The number 5 comes from the four offsets of $x$, $y$, height, width and an extra parameter that detects if the region contains an object or not. YOLO coins it as the “objectness” of the anchor.

\begin{itemize}
\item \textbf{Offset Calculation}: Yolo uses a different formulation to calculate offsets than the Faster RCNN and SSD architectures. The Faster RCNN uses the following formulation:

$$
t_x = (x_g - x_a)/w_a  \eqno{(1)}
$$
$$
t_y = (y_g - y_a)/h_a  \eqno{(2)}
$$
$$
t_w = log(w_g/w_a)  \eqno{(3)}
$$
$$
t_h = log(h_g/h_a)  \eqno{(4)}
$$

This formulation worked well, but the authors of YOLO point out that this formulation is unconstrained. The offsets can be predicted in such a way that they can modify the anchor to lie \textit{anywhere} in the image. Using this formulation, training took a long time for the model to start predicting sensible offsets. YOLO hypothesized that this was not needed as an anchor for a particular position would only be responsible for modifying its structure \textit{around that position} and not to any location in the entire image.

YOLO introduced a new formulation for offsets that constrained the predictions of these offsets to near the anchor box.  
The new formulation modified the above objective by training the network to predict these 5 values.
$$
b_x = \sigma(t_x) + c_x \eqno{(21)}
$$
$$
b_y = \sigma(t_y) + c_y \eqno{(22)}
$$
$$
b_w = w_a*e^{t_w} \eqno{(23)}
$$
$$
b_h = h_a*e^{t_h} \eqno{(24)}
$$
$$
b_o = \sigma( p_o ) \eqno{(25)}
$$

Where $b_x$, $b_y$, $b_w$ , $b_h$, $b_x$ $b_o$ are the target x,y coordinates, width, height and objectness. The value of $p_o$ represents the prediction for objectness. The values for $c_x$ and $c_y$ represent the offsets of the cell in the feature map for the x and y axis. This new formulation constrains the prediction to around the anchor box and is reported to decrease training time.

\end{itemize}

\begin{figure*}[htb]
\centering
\includegraphics[width=0.95\textwidth]{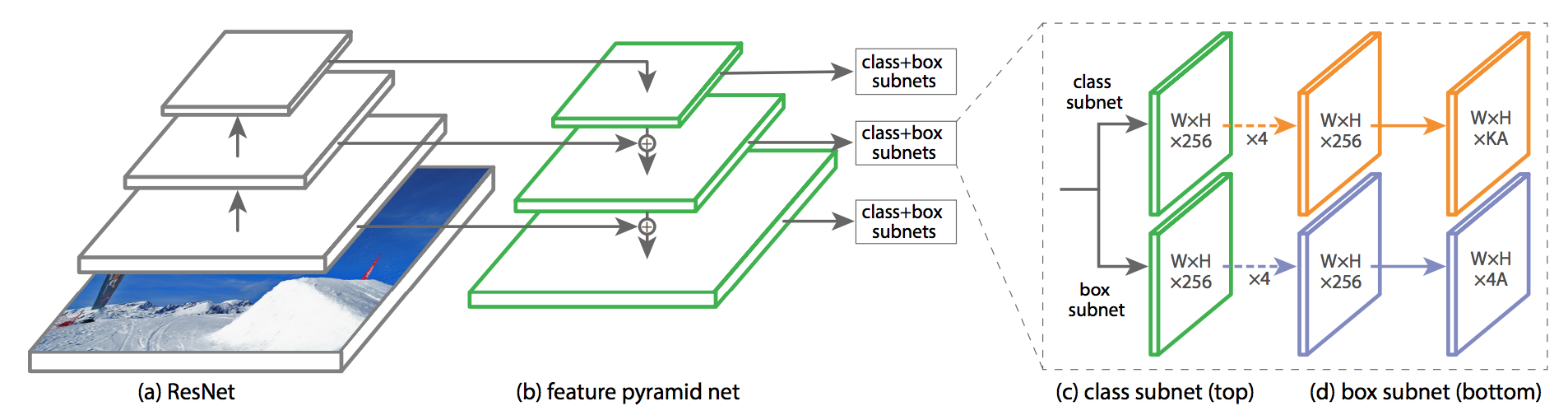}
\label{fig:model8}
\caption{Retina Net}
\end{figure*} 
\section{Retina Net}
\label{sec:retinanet}

The RetinaNet is a single step object detector which boasts the state of the art results at this point in time by introducing a novel loss function \cite{c15}. This model represents the first instance where one step detectors have surpassed two step detectors in accuracy while retaining superior speed.

The authors realized that the reason why one step detectors have lagged behind 2 step detectors in accuracy was an implicit class imbalance problem that was encountered while training. The RetinaNet sought to solve this problem by introducing a loss function coined Focal Loss.

\subsection{Class Imbalance}

Class imbalance occurs when the types of training examples are not equal in number. In the case of object detection, single step detectors suffer from an extreme foreground/background imbalance, with the data heavily biased towards background examples.

Class imbalance occurs because a single step detector densely samples regions from all over the image. This leads to a high majority of regions belonging to the background class. The two step detectors avoid this problem by using an attention mechanism (RPN) which focuses the network to train on a small set of examples. SSD \cite{c5} tries to solve this problem by using techniques like a fixed foreground-background ratio of 1:3, online hard example mining \cite{c8}\cite{c21}\cite{c13} or bootstrapping \cite{c19} \cite{c20}. These techniques are performed in single step implementations to maintain a manageable balance between foreground and background examples. However, they are inefficient as even after applying these techniques, the training data is still dominated by easily classified background examples.

Retina Net proposes a dynamic loss function which down weights the loss contributed by easily classified examples. The scaling factor decays to zero when the confidence in predicting a certain class increases. This loss function can automatically down weight the contribution of easy examples during training and rapidly focus the model on hard examples.

As RetinaNet is a single step detector, it consists of only one model, the object detector model.

\subsection{The Object Detector Model}

RetinaNet uses a simple object detector by combining the best practices gained from previous research.

\begin{itemize}

\item \textbf{Input}: An input image is fed in as the input to the model.

\item \textbf{Targets}: The network uses the concept of anchors to predict regions. As an FPN is integrated with the model, the anchor sizes do not need to account for different scales as that is handled by the multiple feature maps. Each level on the feature pyramid uses 9 anchor shapes at each location. The original set of three aspect ratios ${ {1:1, 1:2, 2:1} }$ have been augmented by the factors of ${ {1, 2^1/3, 2^2/3 } }$ for a more diverse selection of bounding box shapes. Similar to the RPN, each anchor predicts a class probability out of a set of $K$ object classes (including the background class) and 4 bounding box offsets.

Targets for the network are calculated for each anchor as follows. The anchor is assigned the ground truth class if the IOU of the ground truth box and the anchor $\geq$ 0.5. A background class is assigned if the IOU $\leq$ 0.4 and if the 0.4 $\leq$ IOU $\leq$ 0.5, the anchor is ignored during training. Box regression targets are computed by calculating the offsets between each anchor and its assigned ground truth box using the same method used by the Faster RCNN. No targets are calculated for the anchors belonging to the background class, as the model is not trained to predict offsets for a background region.


\begin{table*}[ht]
\caption{Object Detection Comparison Table}
\label{Object Detectors Comparison}
\begin{center}
\begin{tabular}{|c|c||c|c|c|c|c|c|}
\hline
\textbf{Object Detector Type} & \textbf{backbone} & \textbf{$AP$} & \textbf{$AP_{50}$} & \textbf{$AP_{75}$}  & \textbf{$AP_S$} & \textbf{$AP_M$} &  \textbf{$AP_L$}\\ [.1em]
\hline 
~Faster R-CNN+++ \cite{c6} & ResNet-101-C4
  & 34.9 & 55.7 & 37.4 & 15.6 & 38.7 & 50.9\\
\hline
~Faster R-CNN w FPN \cite{c6}& ResNet-101-FPN & 36.2 & 59.1 & 39.0 & 18.2 & 39.0 & 48.2\\
\hline
~Faster R-CNN by G-RMI \cite{c6} & Inception-ResNet-v2 \cite{c3}
  & 34.7 & 55.5 & 36.7 & 13.5 & 38.1 & 52.0\\
\hline
~Faster R-CNN w TDM \cite{c6} & Inception-ResNet-v2-TDM & 36.8 & 57.7 & 39.2 & 16.2 & 39.8 & \textbf{52.1} \\
\hline 
 ~YOLOv2 \cite{c11} & DarkNet-19 \cite{c11}
  & 21.6 & 44.0 & 19.2 & 5.0 & 22.4 & 35.5 \\
\hline
~SSD513 \cite{c5,c4} & ResNet-101-SSD
  & 31.2 & 50.4 & 33.3 & 10.2 & 34.5 & 49.8 \\
\hline
 ~DSSD513 \cite{c18,c4} & ResNet-101-DSSD
  & 33.2 & 53.3 & 35.2 & 13.0 & 35.4 & 51.1 \\
\hline
 ~\textbf{RetinaNet}\cite{c16,c4} & ResNet-101-FPN
  & 39.1 & 59.1 & 42.3 & 21.8 & 42.7 & 50.2 \\
\hline
 ~\textbf{RetinaNet}\cite{c16,c17} & ResNeXt-101-FPN
  & \textbf{40.8} & \textbf{61.1} & \textbf{44.1} & \textbf{24.1} & \textbf{44.2} & 51.2 \\
\hline
\end{tabular}
\end{center}
\label{tab:final_bbox}\vspace{-3mm}
\end{table*}


\item \textbf{Architecture}: The detector uses a single unified network composed of a backbone network and two task specific subnetworks. The first subnetwork predicts the class of the region and the second subnetwork predicts the coordinate offsets. The architecture is similar to an RPN augmented by an FPN.

In the paper, the authors use a Resnet for the convolutional base which is augmented by an FPN to create a rich feature pyramid of the image. The classification and the regression subnets are quite similar in structure. Each pyramid level is attached with these subnetworks, weights of the heads are shared across all levels. The architecture of the classification subnet consists of a small FCN consisting of 4 convolutional layers of filter size 3 by 3. Each convolutional layer has a relu \cite{c49} activation function attached to it and maintains the same channel size as the input feature map. Finally, sigmoid activations are attached to output a feature map of depth $A*K$. The value for $A = 9$ and it represents the number of aspect ratios per anchor, $K$ represents the number of object classes. The box regression subnet is identical to the classification subnet except for the last layer. The last layer has the depth of $4*A$. The 4 indicates the width, height and x and y coordinate offsets. The authors claim that a class agnostic box regressor like the one described above is equally accurate inspite of having having fewer parameters.

\item \textbf{Loss}: The paper pioneered a new loss function called the focal loss. This loss is used to train the entire network and is the central innovation of RetinaNet. It is due to this loss that the network is able to achieve state of the art accuracies while still retaining real time speeds. Before describing the loss, a brief word on backpropogation. Backpropogation is the algorithm through which neural networks learn. It tweaks the weights of the network slightly such that the loss is minimized. Hence, the loss controls the amount by which gradients are tweaked. A high loss for an object makes the network more sensitive for that object and vice versa. 

The focal loss was introduced to solve the class imbalance problem. Methods to combat class imbalance have been used in the past with the most common being the balanced cross entropy loss. The balanced cross entropy loss function down weights the loss generated by the background class hence reducing it's effect on the parameters of the network. This is done using a hyper parameter called $\alpha$. The balanced cross entropy, $B_c$ is given as follows:

$$ 
B_c = \alpha * c \eqno{(26)}
$$
where $c$ is cross entropy. The value for $\alpha$ is as is for the foreground class and $1-\alpha$ for the background class. The value for $\alpha$ could be the inverse class frequency or can be treated as a hyperparameter to be set during cross validation. It is used to balance between foreground and background. The problem however is that this loss does not differentiate between easy/hard examples although it does balance the importance of positive/negative examples.

The authors discovered that the gradients are mostly dominated by easily classified examples. Hence, they decided to down weight the loss for a prediction of high confidence. This allows the network to focus on hard examples and learn how to classify them. To achieve this, the authors combined the balanced cross entropy loss and this discovery of down weighting the easily classified examples to form the focal loss $F_L$.

$$
F_L = (1- p)^\gamma * B_c \eqno{(27)}
$$

where $(1- p)^\gamma$ is called the modulating factor of $F_L$. The modulating factor down weights the effect of the loss if the examples are easy to predict. The factor of $\gamma$ adds an exponential factor to the scale, it is also called the scaling factor and is generally set to 2. For example, the focal loss generated by a example predicted to be of 0.9 confidence, will be down weighted by a factor of 100 while a example predicted to be 0.99 will be down weighted by a factor of 1000.

\item \textbf{Training and Inference}: Training is performed using Stochastic Gradient Descent (SGD), using a initial learning rate of 0.01, which is divided by 10 after 60k examples and and again after 80k examples. SGD is initialized with a weight decay of 0.0001 and a momentum of 0.9. Horizontal image flipping is the only data augmentation technique used. During the start of training, the focal loss fails to converge and diverges early in training. To combat this, in the beginning the network predicts a probability of 0.01 for each foreground class, to stabilize training.

Inference is performed by running the image through the network. Only the top 1k predictions are taken at each feature level after thresholding the detector confidence at 0.05. Non Maximum Suppression (NMS) \cite{c1} is performed using a threshold of 0.5 and the boxes are overlaid on the image to form the final output. This technique is seen to improve training stability for both cross entropy and focal loss in the case of heavy class imbalance.

\item \textbf{Salient Points}

The RetinaNet boasts state of the art accuracy presently and operates at around 60 FPS. It’s use of the focal loss allows it to have a simple design that is easy to implement. Table 1 compares and contrasts different object detectors.
\end{itemize} 

\section{Metrics To Evaluate Object Recognition Models}
\label{sec:metrics}

There have been several metrics that the research community uses to evaluate object detection models. The most important being the Mixed Average Precision and Average Precision metrics.

\subsection{Precision \& Recall}

$$
T_P = True Positive\eqno{(28)}
$$
$$
T_N = True Negative\eqno{(29)}
$$
$$
F_P = False Positive\eqno{(30)}
$$
$$
F_N = False Negative\eqno{(31)}
$$
$$
P = T_P / (T_P+F_P)\eqno{(32)}
$$
$$
R = T_P / (T_P+F_N) \eqno{(33)}
$$

where Precision is $P$ and Recall is $R$

Intuitively, precision measures how accurate the predictions are and recall measures the quality of the positive predictions made by the model. There is generally a trade off while constructing machine learning models with the ideal scenario being a model with a high precision and a high recall. However, some use cases call for greater precision than recall or vice versa.

\subsection{Average Precision}

Average precision is calculated by taking the top 11 predictions made by the model for an object. For these 11 predictions, the precision and recall for each prediction is measured given that $T_P$ is known for the experiment. The prediction is said to be correct if it's IOU is above a certain threshold. These IOU values generally vary between 0.5 and 0.95.

Once precision and recall are calculated each of these 11 steps, the maximum precision is calculated for the recall values that range from 0 to 1 with a step size of 0.1. Using these values, the Average Precision is calculated by taking the average over all these maximum precision values.

The following formula is used to calculate the average precision of the model.

$$
AP_r(i) = \max_{i \leq j}(P_j) \eqno{(34)}
$$

$$
AP = (1/11)*\sum_{r=0}^{1}(AP_r(i)) \eqno{(35)}
$$

where $AP$ is Average Precision. It is averaged over 11 quantities because the step size of $i$ is 0.1.

\subsection{Mean Average Precision (mAP)}

The Mean Average Precision is one of the more popular metrics used to judge object detectors. In recent papers, object detectors are compared via their $mAP$ score. 
Unfortunately the metric  has been used to varying meanings. The YOLO paper as well as the PASCAL VOC \cite{c36} dataset details $mAP$ to be the same quantity as AP. The COCO dataset \cite{c37} however, uses a modification to this metric called the Mixed Average Metric. The AP values for different IOU values are calculated for the COCO $mAP$. The IOU values range from 0.5 to 0.95 with a step size of 0.05. These AP values are then averaged over to get the COCO Mixed Average Precision metric. The YOLO model reports better accuracies in the simple AP of 0.5  metric but not with the $mAP$ metric. This paper treats the COCO $mAP$ as the Mixed Average Precision.

\section{Convolutional Bases}
\label{sec:conv-bases}

All modern object detectors have a convolutional base. This base is responsible for creating a feature map that is embedded with salient information about the image. The accuracy for the object detector is highly related to how well the convolutional base can capture meaningful information about the image. The base takes the image through a series of convolutions that make the image smaller and deeper. This process allows the network to make sense of the various shapes in the image. 

Convolutional networks form the backbone of most modern computer vision models. A lot of convolutional networks with different architectures have come out in the past few years. They are roughly judged on three factors namely accuracy, speed and memory. 

Convolutional bases are selected according to the use case. For example, object detector’s on the phone will require the base to be small and fast. Alternatively, larger bases will be used by the powerful GPU's on the cloud.
A lot of research has gone into making these convolutional nets faster and more accurate. A few popular bases are described in the coming section. Bigger nets have led in accuracy however, advancements have been made to compress and optimize neural networks with a minimal tradeoff on accuracy.

\subsection{Resnet}

Resnet \cite{c4} is an extremely popular convolutional network. It popularized  the concept of skip connections in convolutional networks. Skip connections add or concatenate  features of the previous layer to the current layer. This leads to the network propagating gradients much more effectively during backpropogation. Resnets were the state of the art at the time they were released and are still quite popular today.

The innovation of introducing skip connections resulted in the training of extremely deep networks without over fitting. Resnets are usually used with powerful GPU’s as their processing takes substantially more time on a CPU. These networks are a good choice for a convolutional base on a powerful cloud server.

\subsection{Resnext}

Resnext \cite{c17} networks are an evolution to  the Resnet. They introduce a new concept called grouped convolutions.
Traditional convolutions operate in three dimensions- width, height and depth. Grouped convolutions introduce a new dimension called cardinality. 

\textit{Cardinality} espouses dividing a task into $n$ number of smaller subtasks. Each block of the network goes through a 1 by 1 convolution similar to the Resnet, to reduce dimensionality. The next step is slightly different. Instead of running the map through a 3 by 3 convolutional layer, the network splits the $m$ channels (where $m$ is the depth of the feature map after the 1 by 1 convolution) into groups of $n$, where $n$ is the cardinality. A 3 by 3 convolution is performed for each of these $n$ groups (with $m/n$ channels each) after which, the $n$ groups are concatenated together. After concatenation, this aggregation undergoes another 1 by 1 convolution layer to adjust the channel size.
Similar to the Resnet, a skip connection is added to this result.

The usage of grouped convolutions in Resnext networks led to a better classification accuracy while still maintaining the speed of a Resnet network. They are indeed the next version of Resnets.

\subsection{MobileNets}

These networks are a series of convolutional networks made with speed and memory efficiency in mind. Like the name suggests, the MobileNet \cite{c14}\cite{c16} class of networks are used on low powered devices like smart phones and embedded systems.

MobileNets introduce depth wise separable convolutions that lead to a major loss in floating point operations while still retaining accuracy. The traditional procedure of converting 16 channels to 32 channels is to do it in one go. The floating point operations are $(w*h*3*3*16*32)$ for a 3 by 3 convolution.

Depth wise convolutions make the feature map go through a 3 by 3 convolution by not merging anything, resulting in 16 feature maps. To these 16 feature maps, a 1 by 1 filter with 32 channels is applied, resulting in 32 feature maps. Hence, the total computation is $(3*3*16 + 16*32*1*1)$ , which is far less than the previous approach.

Depth wise separable convolutions form the backbone of MobileNets and have led to a speedup in computing the feature map without sacrificing the overall quality. The next generation of MobileNets have been released recently are are aptly named MobileNetsV2 \cite{c16}. Apart from using depth wise convolutions they add two new architectural modifications- a linear bottleneck and skip connections on these linear bottlenecks. Linear bottleneck layers are layers that compress the number of channels from the previous layer. This compression has been experimentally proven to help, they make MobileNetsV2 smaller but just as accurate. The other innovation in MobileNetV2 \cite{c16} over the V1 \cite{c14} is skip connections which were popularized by the Resnet. Studies have shown that MobileNetsV2 are 30-40\% faster then the previous iteration.

Papers have coined object detection systems using MobileNets as the SSDLite Framework. The SSDLite outperforms the Yolo architecture by being 20\% more efficient and 10\% smaller. Networks like the Effnet \cite{c39} and Shufflenet \cite{c40} are also some examples of promising steps in that direction.

\section{Training}
\label{sec:training}

There have been significant innovations in training neural networks in the past few years. This section provides a brief overview of some new promising techniques.

\subsection{Superconvergence}

Superconvergence is observed when a model is trained to the same level of accuracy in exponentially lesser time than the usual way using the same hardware. An example of Superconvergence is observed when training the Cifar10 network to an accuracy of 94\% in around 70 epochs, compared to the original paper which took around 700 epochs. Superconvergence is possible through the use of the 1 Cycle Policy to train neural networks.

To use the 1 Cycle Policy \cite{c27}, the concept of a \textit{learning rate finder} needs to be explained. The learning rate finder seeks to find the highest learning rate to train the model without divergence. It is helpful as one can now be sure that the training of the model is performed at the fastest rate possible. To implement the learning rate finder, the following steps are followed:

\begin{itemize}

\item Start with an extremely small Learning Rate (around 1e-8) and increase the Learning Rate linearly.
\item Plot the loss at each step of the Learning Rate.
\item Stop the learning rate finder when the loss stops decreasing and starts increasing.
\end{itemize}

After observing the graph, a value for the Learning Rate is decided upon by taking the maximum value of the Learning Rate when the loss is still decreasing. Now once the optimal Learning Rate (L) is discovered, the 1 Cycle Policy can be used. 
 \begin{figure}[thpb]
      \centering
      \includegraphics[scale=0.5]{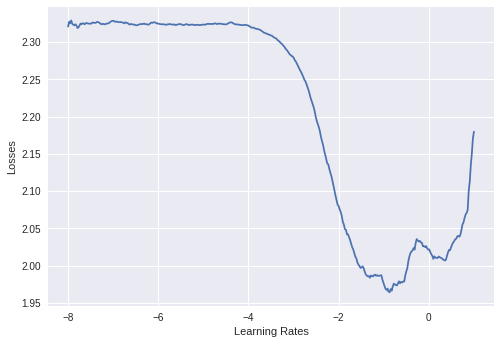}
      \caption{Learning Rate Finder}
      \label{fig:model9}
   \end{figure}

The 1 Cycle Policy \cite{c27} \cite{c46} states that to train the model, the following steps should be observed:

\begin{itemize}
\item Start with a Learning Rate L/10 than the learning rate found out by the Learning Rate finder.
\item Train the model, while increasing the Learning Rate linearly to L after each epoch.
\item After reaching L, start decreasing the Learning Rate back till L/10.
\end{itemize}

 \begin{figure}[thpb]
      \centering
      \includegraphics[scale=0.5]{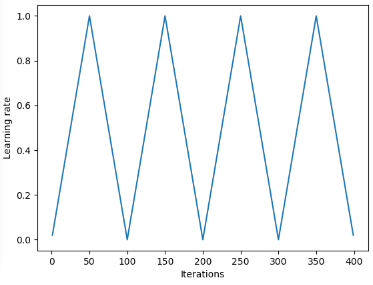}
      \caption{Feature Pyramid Network}
      \label{fig:model10}
   \end{figure}

\subsection{Distributed Training}

Training huge models on a single machine is not feasible. Nowadays, training is distributed on various machines. Distribution aids parallelism which leads to substantial improvements in training time. Distributed Training \cite{c42}\cite{c43}\cite{c44} has led to datasets like Imagenet being trained in as little a time as \textit{4 minutes}. This has been made possible by using techniques like  Layer-wise Adaptive
Rate Scaling (LARS). The major intuition is that all the layers of a neural network shouldn't be trained at the same rate. The initial layers should be trained faster than the last layers at the beginning of training, with this being reversed when the model has been training for some time. Using adaptive learning rates has also led to substantial improvements in the training process. The most popular way of training networks presently is to use a small learning rate to \textit{warm up} the network after which higher learning rates are used with a decay policy.

As of this time, techniques like LARS and the 1 Cycle Policy haven't been used to train object detection algorithms. A paper exploring these new training techniques for the use case of object detection would be quite useful.

\section{Future Work}
\label{sec:future}
Object Detection has reached a mature stage in it’s development. The algorithms described in this paper boast state of the art accuracy that match human capability in most cases. 
Although like all neural networks, they are susceptible to adversarial examples. Better explainability and interpretability \cite{c33} are open research topics as of now. 

Techniques to make models faster and less resource intensive can be explored. Reducing time to train these models is another research avenue. Applying new techniques like super convergence \cite{c27}, cyclic learning rates \cite{c23} and SGDR \cite{c28} to these models, could reveal new state of the art training times. Apart from reducing training time, reducing inference time can also be explored by using quantization \cite{c24}\cite{c25} techniques and experimenting with new architectures. Automated architecture search is a promising step in that direction.

Neural Architecture Search (NAS) \cite{c29} \cite{c30} \cite{c31} tries out various combinations of architectures to get the best architecture for a task. NASNets have achieved state of the art results in various Computer Vision Tasks and exploring NASNets for the object detection problem could reveal some new insights. A caveat of NASNets is that they take an extremely long time to discover these architectures, hence, a faster search would need to be employed to get results in a meaningful time frame.

Weakly supervised training techniques \cite{c32} to train models in the absence of labels is another research direction that holds promise. In the real world, good object detection datasets are rare. Papers combining unlabeled data to train detectors have been coming out recently. Techniques such as using different transformations of unlabeled data to get automatic labels have been researched. These automatically generated labels have been trained along with known labels to get state of the art results in object detection. Some variations to this technique can be explored such as using saliency maps to aggregate predictions, which could prove to generate better quality labels.
The deep learning field is growing rapidly or rather, exploding. There shall be consistent improvements to the above techniques in the coming years.

\section*{Conclusion}

In conclusion, the paper has surveyed various object detection Deep Learning models and techniques, of which the Retina Net is noted as the best till date. The paper has also explored techniques to make networks portable through the use of lighter convolutional bases. In closing, various training techniques that make these networks easier to train and converge have also been surveyed. Techniques such as Cyclic Learning Rates, Stochastic Weight Averaging \cite{c45} and Super Convergence, will lead to better training times for a single machine. For training in a distributed setting, techniques such as LARS \cite{c43} and Linear batch size scaling \cite{c44}\cite{c42} have proven to give substantial efficiencies.

\end{document}